\documentclass[runningheads]{llncs}
\usepackage[T1]{fontenc}
\usepackage{graphicx}
\usepackage[misc]{ifsym}
\usepackage{wrapfig}

\begin{document}
\title{Robotic Calibration Based on Haptic Feedback Improves Sim-to-Real Transfer}
\titlerunning{Robotic calibration for sim-to-real transfer}
%
\author{Juraj Gavura\inst{1} \and
Michal Vavre\v{c}ka\inst{1} 
\and
Igor Farka\v{s}\inst{1}
\and
Connor G\"ade\inst{2}
}
\authorrunning{J. Gavura et al.}
%
\institute{Department of Applied Informatics, Comenius University Bratislava, Slovakia \\ 
\email{gavura4@uniba.sk, michal.vavrecka@cvut.cz, farkas@fmph.uniba.sk} 
\and
Department of Informatics, University of Hamburg, Germany\\
\email{connor.gaede@uni-hamburg.de}
}
%
\maketitle  
\begin{abstract}
When inverse kinematics (IK) is adopted to control robotic arms in manipulation tasks, there is often a discrepancy between the end effector (EE) position of the robot model in the simulator and the physical EE in reality. In most robotic scenarios with sim-to-real transfer, we have information about joint positions in both simulation and reality, but the EE position is only available in simulation. We developed a novel method to overcome this difficulty based on haptic feedback calibration, using a touchscreen in front of the robot that provides information on the EE position in the real environment. During the calibration procedure, the robot touches specific points on the screen, and the information is stored. In the next stage, we build a transformation function from the data based on linear transformation and neural networks that is capable of outputting all missing variables from any partial input (simulated/real joint/EE position). 
Our results demonstrate that a fully nonlinear neural network model performs best, significantly reducing positioning errors.

\keywords{Calibration  \and Humanoid robot \and Inverse kinematics \and Haptic feedback}
\end{abstract}
\section{Introduction}


Recent advances in robotics have significantly improved the capabilities of industrial and humanoid robots, enabling them to perform complex manipulation and interaction tasks with increasing autonomy and dexterity. Modern robotic systems integrate advanced perception techniques, such as deep learning-based vision \cite{levine2018}, tactile sensing \cite{kappassov2015}, and proprioception \cite{hogan2022}, allowing for more adaptive and precise control.
However, to fully utilize their capabilities, robots must precisely control their actuators within their workspace to ensure accurate positioning of the end effector (EE). Small errors in joint actuation can lead to significant deviations in the EE position, affecting performance in tasks such as grasping objects or using tools. This need for precision highlights the importance of robust calibration and compensation techniques.

A significant challenge in robotic manipulation arises when using inverse kinematics (IK) solvers, as the model used in the simulator often fails to precisely match the real-world structure of the robot. In particular, the predicted position of the EE in the simulator may not align with its actual position in the real environment. This discrepancy is typically caused by inaccuracies in the robot's model, such as imperfections in the actuators, sensors, or the simulated physical properties that do not account for real-world dynamics. As a result, even when the IK solver calculates the correct joint angles for a desired EE position, the actual EE may not reach the target location. To address this issue, several compensation techniques have been proposed, including model calibration, sensor fusion, and machine learning-based methods.

In this paper, we propose a novel method to address the discrepancy between simulated and real-world EE positions in robotic systems. Our approach is based on a calibration procedure in which the robot touches specific points on a touchscreen, allowing us to gather data that map the simulated coordinates to real-world coordinates. Using this technique, a transformation function is calculated to bridge the gap between the two coordinate systems. Compared to other techniques, our method offers a low-cost, highly precise solution that does not require specialized hardware. This makes it adaptable to other robotic systems.

\section{Related work}

Several studies have explored approaches to mitigate the sim-to-real gap in robotics, highlighting the effectiveness of integrating simulated data with additional sensory feedback to reduce discrepancies between model predictions and real-world outcomes \cite{alquareshi2020}. Methods for compensating EE errors can be categorized based on how the real EE is measured, ranging from the absence of direct measurements to the use of motion capture, cameras, haptic feedback, or sensor fusion, and the type of compensation algorithm employed, such as conventional analytical methods or neural networks. Previous work combines these strategies to enhance accuracy. In \cite{liu2022}, a two-stage framework was introduced that integrates an enhanced Newton-Raphson method with a ResNet model to refine the mapping of simulated joint angles to real-world EE positions. In \cite{vincente2014}, an online kinematic adaptation method for humanoid robot arms was proposed, utilizing both visual and proprioceptive sensors. The approach employs stereovision-based hand pose estimation, followed by a particle-based optimization to correct kinematic calibration errors. In addition, Bayesian methods have been applied for hand-eye kinematics calibration in humanoid robots, optimizing Denavit-Hartenberg parameters and joint gear reductions to improve accuracy \cite{hubert2012}.

Deep neural networks have been used to facilitate sim-to-real transfer by estimating arm joint configurations from stereo images, allowing precise reach without manual fine-tuning \cite{nguyen2018}. 
This approach was tested in the iCub robot, where a visuomotor predictor provided accurate joint estimates to quantify systematic errors in the robot joint measurements. In the subsequent stage, an automatic calibrator was implemented to compensate for these errors, with the potential to integrate both predictor and calibrator into a unified system. More recently, Gäde et al. \cite{gade2024} proposed an end-to-end visuomotor architecture that incorporates domain adaptation and inherent IK. By embedding domain adaptation within their model, they enabled continuous calibration, which was validated through sim-to-real grasping experiments with the NICO humanoid robot. Their approach significantly improved accuracy, achieving an 80.30 \% success rate.

If visual feedback lacks sufficient precision, a secondary robot or motion tracking system can be used to provide ground truth data for the EE position. 
An industrial Kuka manipulator was employed to calibrate the leg joints of the AR601M humanoid robot \cite{khusainov2017}. Using the precise EE positioning of the Kuka arm, they achieved a highly accurate joint offset calibration for humanoid robots. When ground truth data on EE position is available, calibration can also be extended to robots with elastic components and joints. 
Advanced kinematic models were introduced to account for joint and transverse elasticities, reducing absolute position errors in humanoid upper bodies from 21 to 3 mm \cite{tenhumberg2021}.

If a robot is equipped with artificial skin, calibration can be performed using tactile feedback. Inspired by infant behavior, self-touch calibration has been implemented in humanoid robots, leveraging tactile arrays to refine kinematic representations without requiring external measurement systems \cite{roncone2014}. Self-calibration approaches utilizing multiple kinematic chains have demonstrated superior performance in optimizing robot parameters \cite{stepanova2019}. The authors used a simulated humanoid robot iCub with a stereo camera system and EE contact emulation to perform self-touch-based calibration, achieving an EE accuracy of 2 mm after 100 calibration poses. More recently, a multisensory calibration framework was developed,  combining conventional techniques with self-contained calibration strategies tailored for humanoid robots, facilitated by an open source toolbox for implementation and evaluation \cite{rozlivek2021}.

We can summarize that recent visuomotor coordination-based calibration methods often suffer from imprecision due to limitations in visual perception, such as occlusions, lighting conditions, and camera calibration errors. However, methods that rely on external calibration tools, such as motion capture systems or high-precision external sensors, are costly and require additional hardware, making them impractical for many applications. Another approach, self-touch calibration, provides direct feedback by using the robot’s own EE to interact with its body; however, this method requires robotic skin or tactile sensors, which can be expensive and not available on all robots. 

To address these limitations, we propose a novel method that utilizes tactile feedback from a touchscreen as a calibration tool. This approach is more precise than vision-based methods, as it provides direct contact information without relying on external cameras. Additionally, it is significantly cheaper than motion capture systems and does not require robotic skin, making it a cost-effective and widely applicable solution for various robotic platforms.

\section{Materials and methods}

\begin{figure}[t]
\centering
\includegraphics[width=0.7\textwidth]{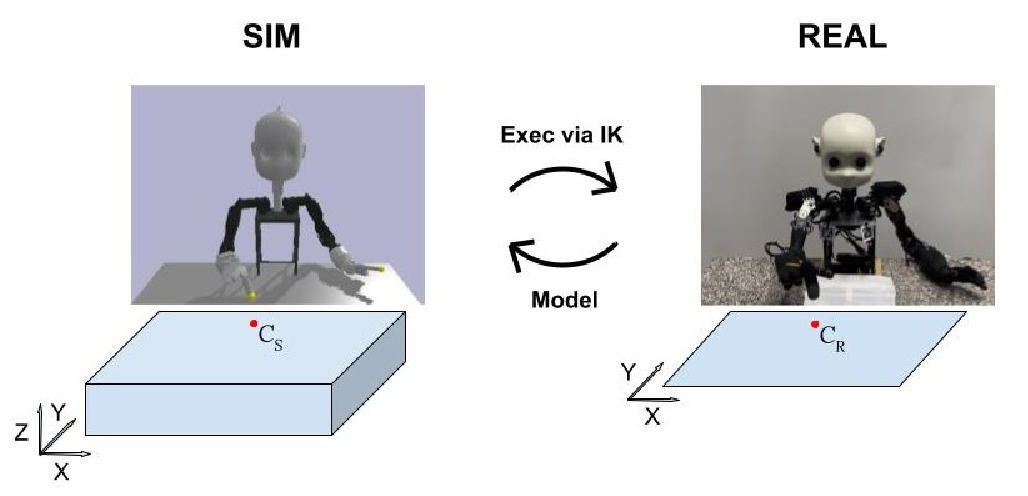}
\vspace*{-5mm}
\caption{Calibration schema of the sim-to-real transfer based on haptic feedback. For any target point on the touchscreen, a model is used to calculate the corresponding 3D position in the simulator. It is used by the IK module to enable the required move.} 
\label{schema}
\end{figure}

Our aim was to ensure that when the robot was directed to a specific target point in Cartesian space, its EE would accurately reach the designated location with minimal deviation.
To achieve this, we established relationships between the real world, with the robot and the touchscreen, and a simulated world.\footnote{This approach may seem contradictory for sim-to-real transfer assuming we start with perfect simulation, which is then transferred to real world. But actually, we need perfect behavior in {\it real} world, so we calibrate the simulation to enable that.}
Creating a perfect match between the Cartesian positions in the two worlds remains a challenge. The overall scheme is shown in Fig.~\ref{schema}.

Both domains use two coordinate systems: joint space and Cartesian space.
In real space, we denote the joint angles by $J_R$, and in simulation $J_S$. 
There, we use $(x,y,z)$ coordinates denoted as $C_S$, which are known at each step. In the real world, the touchscreen provides only 2D feedback, which means that we only have access to $(x,y)$ coordinates denoted as $C_R$.
Given that the touchscreen lies in a fixed plane, we can assume that the $z$-coordinate in reality is always 0 on the touchscreen.

\subsection{Robot and touchscreen}

We conducted all experiments using the humanoid robot NICO, designed for multimodal interaction \cite{hwei2017nico}. 
We used the robot's right arm equipped with 6 motors to control joints and a four-finger child hand with 4 motors. We controlled 7 DoF during the experiment, as touching requires only the index finger.
We used a 1920$\times$1080 px touchscreen fixed in front of the robot as a tactile feedback interface for the calibration procedure. To enable interaction with the screen, the robot's index finger was wrapped in a tactile conductive material, allowing it to make contact and receive feedback on its actual position. The screen with 54$\times$33 cm covers approx.~60\% of the robot's working area on the table. It is fixed at the height of the table, ensuring a stable reference frame and allowing a precise mapping of each pixel to a corresponding position $C_R$. 
Before working with NICO, we assessed the physical limitations of its arm and identified a specific screen region that NICO's right hand could reliably reach. Afterwards, the robot was ready for movement and calibration.

\subsection{NicoIK}

 The first part of our calibration is the publicly available software NicoIK\footnote{https://github.com/Robotics-DAI-FMFI-UK/NicoIK} that can be adapted with modifications for other robots. The software consists of several components. The simulation module is built on the PyBullet physics engine \cite{coumans2021} and uses its IK solver and some functions from myGym \cite{vavrecka2021}. It can calculate IK for any robot with a proper body model. We fine-tuned the NICO URDF file to better match the actual hardware, ensuring precise alignment between the simulated and real robot, which helps reduce final calibration errors. The robot control module is based on the NicoMotion library, which is part of \cite{nico_software} and can be exchanged with other control libraries to adopt a different robot. The NicoIK orchestrates these components and records calibration data. The results were then used as input for the error correction algorithms (see below).

\subsection{Sim-to-real alignment}

The first step of our calibration process was to determine the exact position of the touchscreen within the simulation, ensuring proper alignment of the virtual environment. To achieve this, we manually guided the robot's end-effector to three reachable edge points of the touchscreen in the real world and recorded the corresponding joint angle values of the robot's arm. Using forward kinematics, we then mapped these recorded joint angles to their respective positions in the simulated environment, allowing us to accurately establish the touchscreen's placement in the simulation.

However, three reference points differed not only in their $x$ and $y$-coordinates, but also in the $z$ coordinate (height), resulting in a slightly tilted touchscreen in the simulation. By applying linear interpolation between these points, we could determine any position on the simulated screen. To test the initial transition from simulation to reality, we generated multiple points on the simulated screen, computed the corresponding joint angles using IK, and executed these movements on the real robot. To assess accuracy, we relied on the touchscreen as the only ground truth by having the robot physically touch it. However, this approach did not guarantee precise alignment, revealing nonlinear discrepancies in the $z$-coordinate between the touchscreen positions in simulation and real world.

\begin{figure}
\centering
\includegraphics[width=0.8\textwidth]{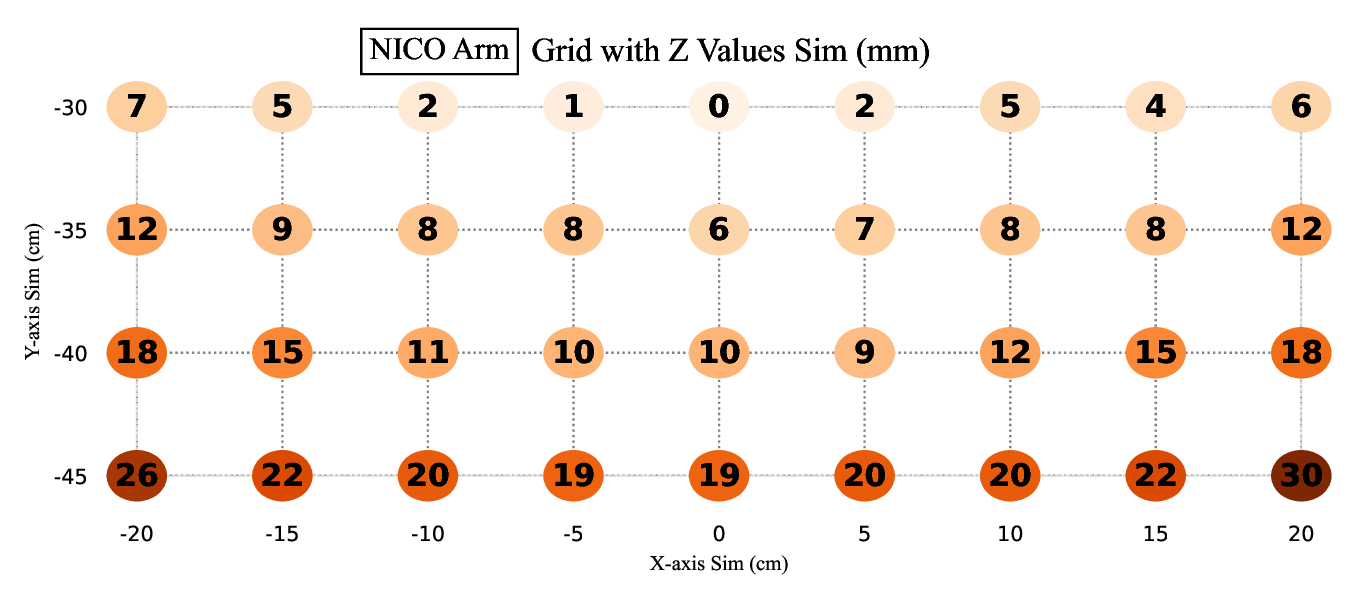}
\vspace*{-4mm}
\caption{Visualisation of manually measured $z$ values (height) of touchscreen in grid-based points in  simulation. Real NICO will touch the screen at point ($x,y$) from the grid if we set the goal $z$ based on the corresponding $z$ value in this grid. We can see that the differences are roughly symmetrical with respect to $x$ axis.} 
\label{grid_z}
\end{figure}

The simplest approach to address this nonlinearity issue was to empirically determine the $z$-coordinate for a set of regularly spaced points across the utilized portion of the touchscreen and then apply linear interpolation using four nearest points (on the grid). To achieve this, we generated a grid of 36 points on the simulated screen. The EE of the robotic arm was then guided to each of these points, gradually decreasing the $z$-coordinate until the robot's fingertip reached contact with the real touchscreen. At the moment of contact, we record the corresponding $z$-coordinate. This process provided us with the measured heights of 36 uniformly distributed points on the touchscreen (see Fig.~\ref{grid_z}). With this dataset, we could now estimate the height of any point in the simulation by linearly interpolating between its nearest measured grid points.

\subsection{Calibration models}

\subsubsection{Piecewise linear model (M1)}
M1 takes a point $C_R$ as input, linearly interpolates its coordinates in simulation using the edge points gathered when locating the touchscreen, and then interpolates the $z$-coordinate  using the measured heights in the grid. This yields $C_S$, linearly mapping the real data to the simulation. This model served as a baseline for testing more advanced models.
For sim-to-real transfer, we used IK to compute $J_S$ corresponding to $C_S$. Once the real robot executed the movements, we recorded the EE touch positions on the real screen ($C_R$) and corresponding $J_R$ after each motion. Based on previous experience,  we tested three movement durations: one, two, and three seconds. For each duration, we collected data from 250 points, randomly distributed, or in specific patterns, and compared them to the simulated values. To achieve the most natural movement, we prioritized motions with a 2-sec duration.

\vspace{-5mm}
\subsubsection{Partially-nonlinear neural network model (M2)}
Data gathered using M1 was used to develop two more complex nonlinear models. Specifically, the data provided insight into where a real EE would move when given a corresponding point in the simulator. This relationship can also be viewed in reverse: To guide the robot's index finger to a specific point $C_R$, we must supply the appropriate point $C_S$. As a result, for 250 real-world points $C_R$($x,y,0$), we identified 250 corresponding points in the simulation $C_S$($x,y,z$). Since the mapping between these points is nonlinear, we required adequate methods to enhance accuracy.
We trained a partially nonlinear model M2, based on a multilayer perceptron with 16 hidden neurons and the ReLU activation function, which takes $C_R(x,y)$ as input and outputs the corresponding $C_S(x,y)$. 
The $z$-coordinate is estimated using the same linear interpolation method as in M1.
The M2 was trained for $200$ epochs with the Adam algorithm using the learning rate of 0.001, $\beta_1=0.9, \beta_2=0.999, \epsilon=10^{-7}$.

\vspace{-5mm}
\subsubsection{Nonlinear neural network model (M3)}
M3 was chosen to be a fully nonlinear approach that, unlike previous models, approximates a 3D point $C_S(x,y,z)$ from a 2D real point $C_R(x,y)$, hence eliminating the need for linear interpolation. M3 also has one hidden layer with ReLU, but with 64 neurons.
Both neural networks were trained using the Adam optimizer minimizing the MSE with all data derived from the baseline M1 model. Then we tested both models M2 and M3, applying the same procedure as M1 but only on 250 randomly distributed points, and compared the results against the simulated values. The hyperparameters of M3 were the same as in M2.

\subsection{Calibration procedure}

In summary, our calibration procedure includes $6$ steps to obtain a model that minimizes the error in the sim-to-real transfer.

\begin{enumerate}
\item Find the operational space of the robot and align it with the haptic space of the touchscreen.
\item Locate the touchscreen in the simulated space.
\item Generate interpolated grid-based targets that lie on the simulated touchscreen and measure the value $z$ for each of these points.
\item Create a data set of 2D interpolated targets covering the entire touchscreen and interpolate the $z$ value for each target using the data from step 3.
\item Navigate a real robot to these targets using inverse kinematics and save the resulting positions of the hits.
\item Use data collected in the previous step to train the neural network that predicts the simulated position $C_S(x,y,z)$ for any real point $C_R(x,y)$.
\end{enumerate}

\begin{figure}[t!]
\includegraphics[width=\textwidth]{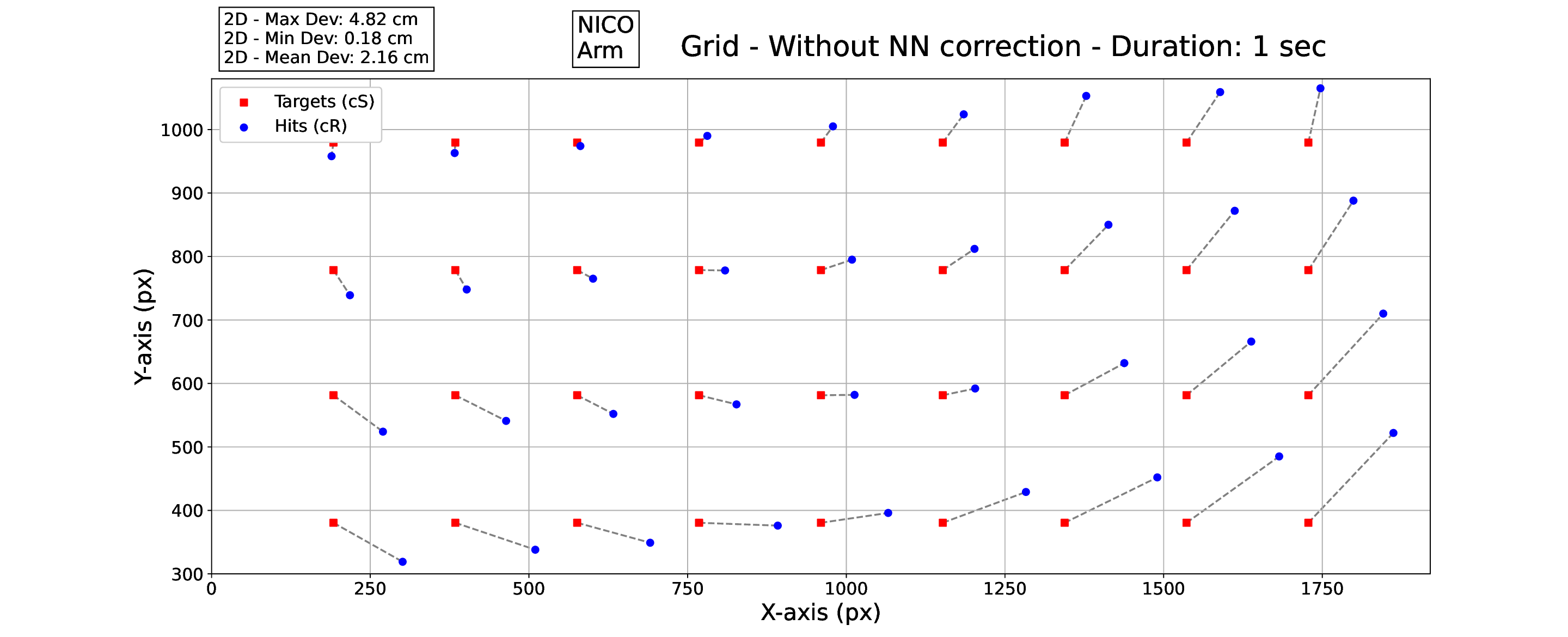}
\includegraphics[width=\textwidth]{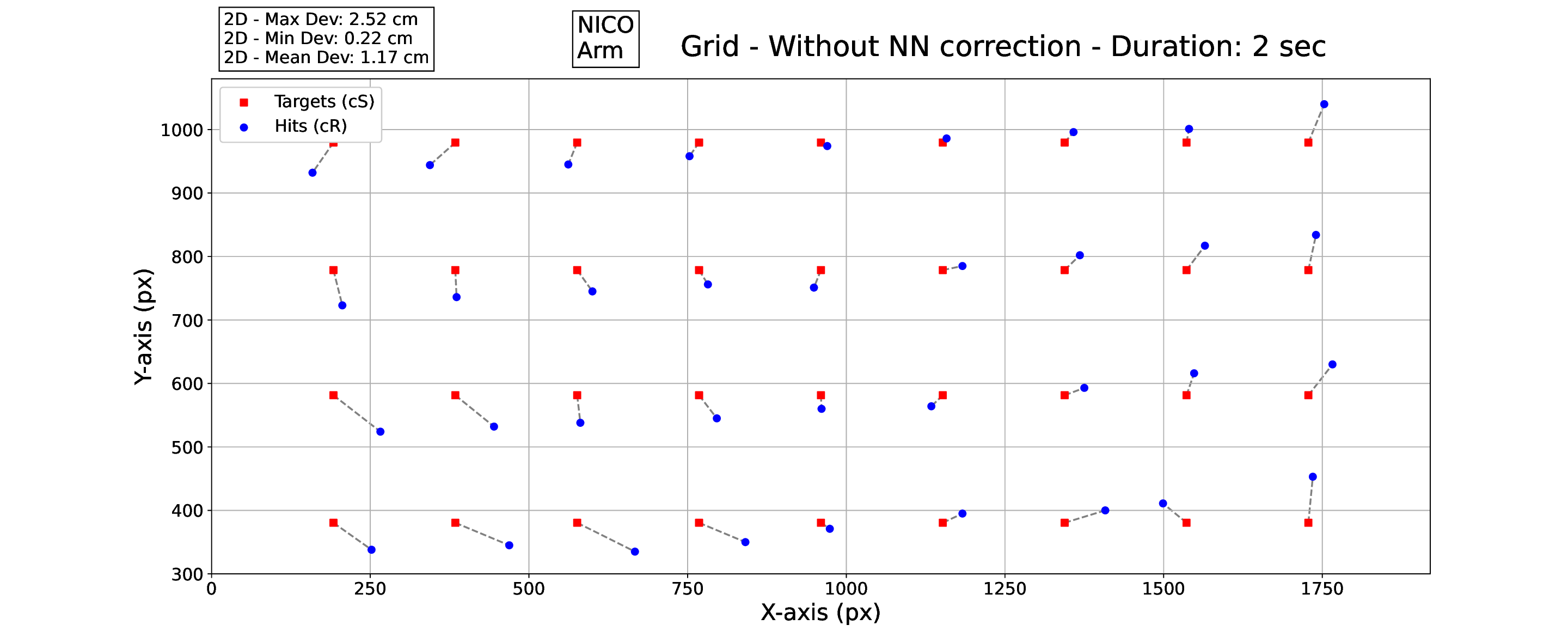}
\caption{Plot of deviations of hits (blue) from grid-based targets (red) in the physical space (on the touchscreen). We used the model M1 and NICO arm movement of 1-sec (top) and 2-sec (bottom) duration. The box "NICO Arm"  denotes the base position.} 
\label{grid_1-2s}
\end{figure}

\vspace*{-1mm}

\section{Results}

\subsection{Model M1 - baseline}

The model M1 serves as the baseline for our research. We applied M1 
with a 1-sec. movement to grid-based points, and plotted the contact points on the screen relative to the intended targets. 
The results in the upper half of Fig.~\ref{grid_1-2s} reveal that the direction of each deviation rotates depending on the target position relative to the arm base. 
In addition, the magnitude of the deviation increases considerably with the target distance from the base (i.e. for more stretched arm).

A similar relationship can be observed in lower half of Fig.~\ref{grid_1-2s}, showing 2-sec movements. By comparison, we see that the average deviation decreased from 2.16 cm (for 1-sec) to 1.17 cm (for 2-sec), demonstrating that slower movements of NICO produce smaller errors. 
Further, we can see that the rotation trend is not as visible here at a 1-sec duration. Despite the slower and more precise movements, the direction of deviation from the target is less regular. 3-sec movements had very similar results to 2-sec movements.

\begin{figure}[t]
\includegraphics[width=\textwidth]{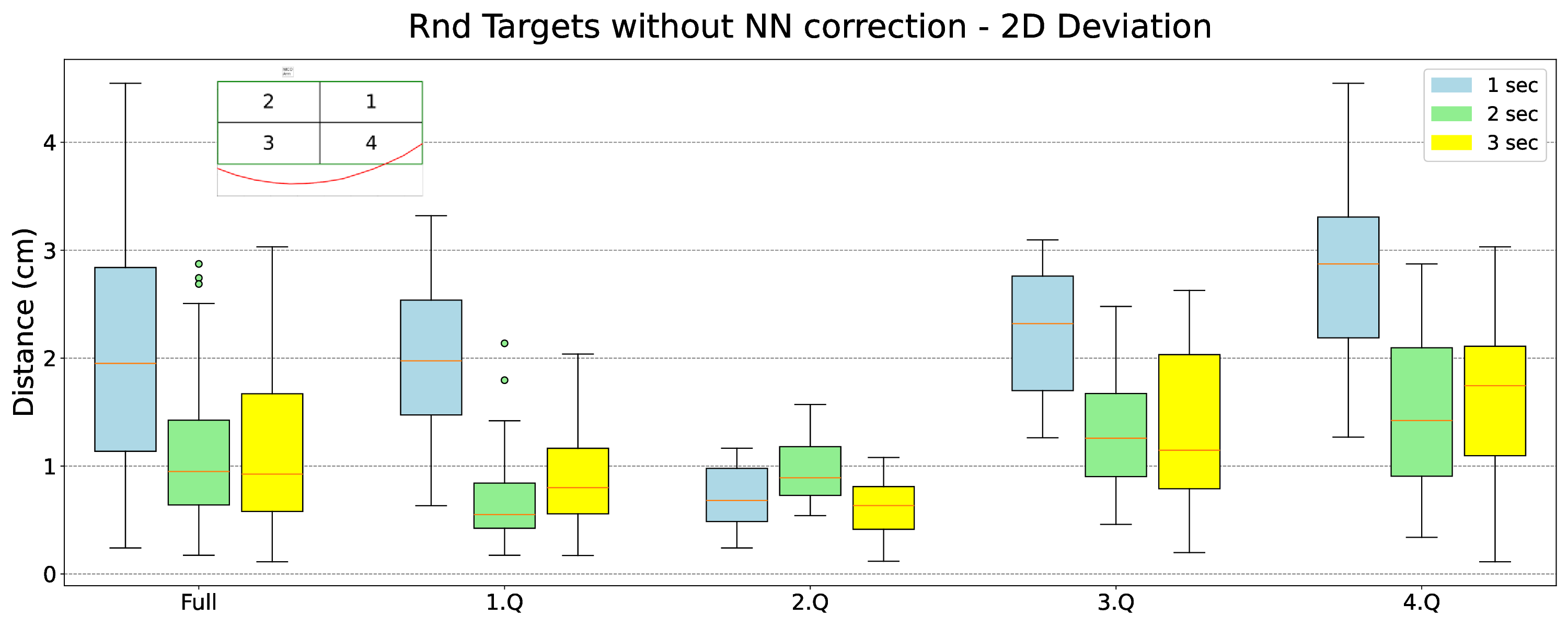}
\caption{Comparison of 2D deviations of hits from random targets within four quadrants for three durations of movement. Targets were calculated using M1. Small inset image top left shows the NICO reach limit (red line) and partitioning of the testing area.} 
\label{M1-Cartes}
\end{figure}

Next, we measured 2D deviations from the target on the touchscreen at three different movement durations for each quadrant, as shown in Fig.~\ref{M1-Cartes}.
We can see that 1-sec movements are on average twice as inaccurate as slower movements, which may be due to overshooting the target. Movements of 2 and 3 seconds reveal similar behavior. We can also see that targets in Q1 and Q2, being closer to the robot base, yield a smaller error compared to more distant Q3 and Q4. 

\begin{figure}[h!]
\includegraphics[width=\textwidth]{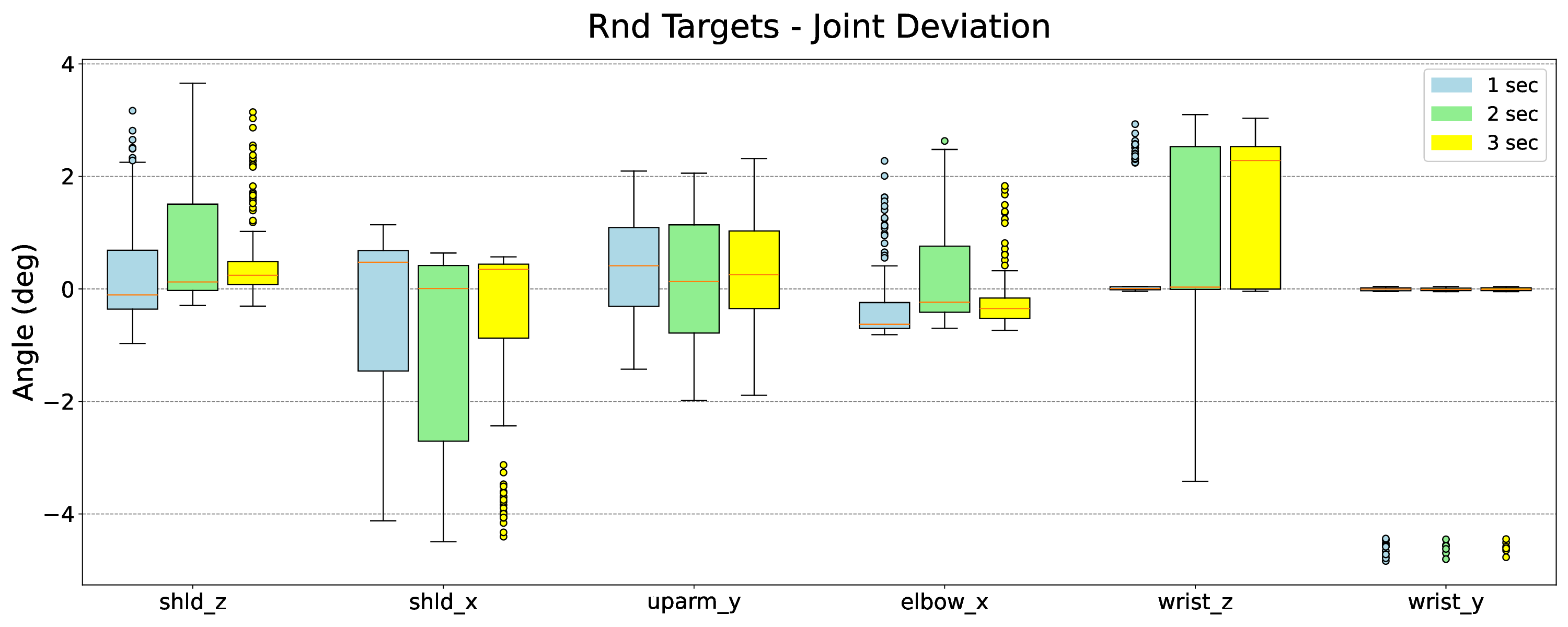}
\caption{Comparison of joint angle deviations for three different durations of movement. Joint deviation is calculated as a difference of joint angles returned from inverse kinematics (angles sent to robot) and joint angles read from the robot after the end of movement (using M1). 
}
\label{M1-joints}
\end{figure}

Fig.~\ref{M1-joints} shows how the joint values of the robotic arm vary from the target positions after execution of the movement and compares similarly three different movement durations. The differences between the movement durations are not large. The deviation of the joint wrist$\_$z, which moves the robot's palm, changes the most as the duration of the movement increases. For 3-sec movements, the median of the measured deviations for this joint reaches more than 2 degrees. In general, the largest deviations are for the joints shld$\_$z and shld$\_$x, which control the shoulder and move and rotate with the whole arm. This is probably due to the fact that the greatest force is applied to these joints.

\subsection{Nonlinear models}

\begin{wrapfigure}{r}{0.45\textwidth}
\vspace*{-5mm}
\includegraphics[width=0.45\textwidth]{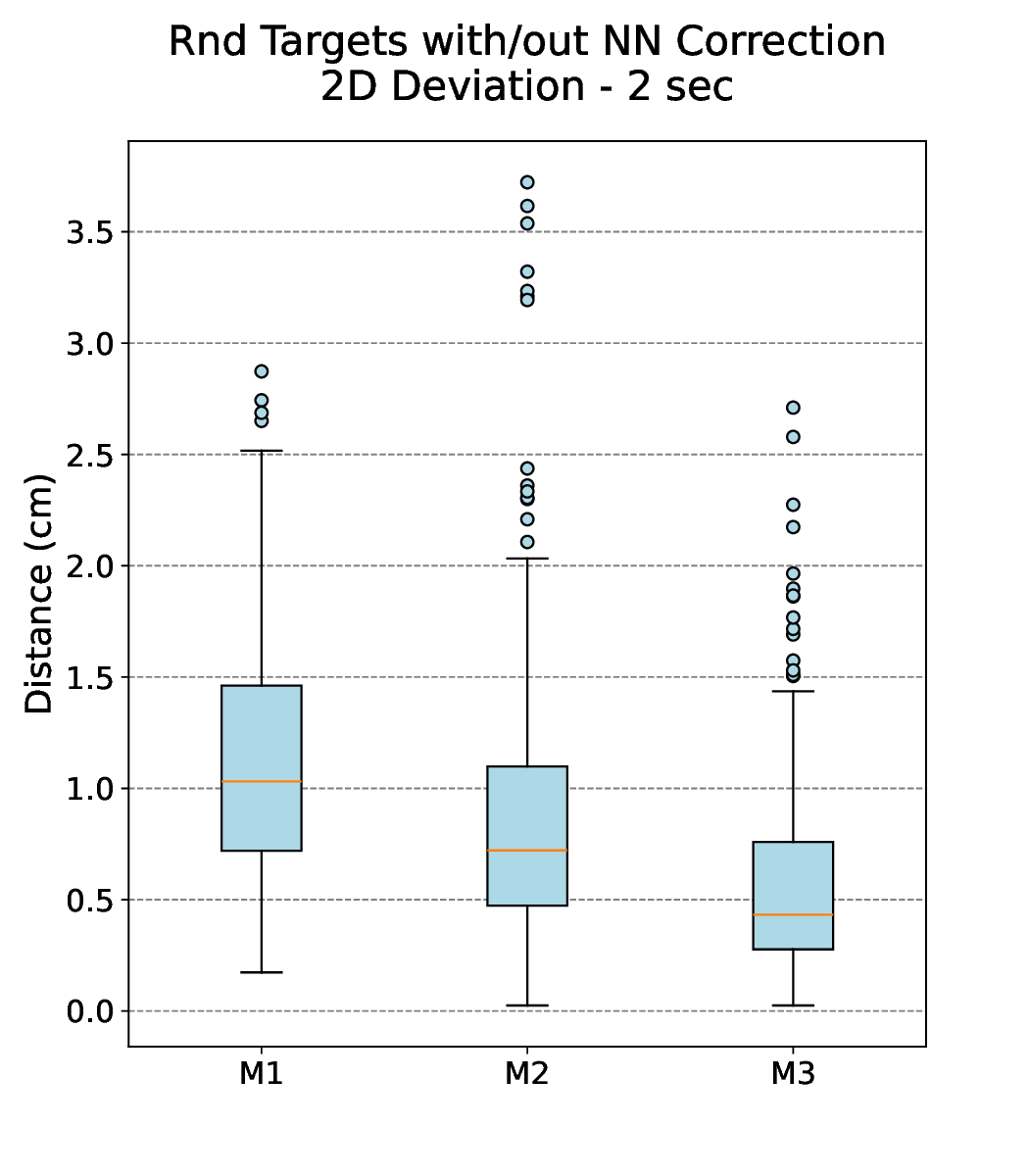}
\vspace*{-10mm}
\caption{Comparison of 2D deviations (Euclidean distances) of individual model predictions for random targets.}
\vspace*{-7mm}
\label{models_boxplot}
\end{wrapfigure}

We implemented nonlinear models (M2 and M3) based on neural networks to solve the sim-to-real transfer and compared them with a linear baseline M1.
The neural networks were tested by cross-validation five times, and achieved high accuracy with MSE $<$ 0.001.
With the data obtained from M2 and M3, we compared their performance against M1. As shown in Fig.~\ref{models_boxplot}, both nonlinear models significantly improve accuracy compared to M1, with M3 performing the best.
This indicates that predicting $x$, $y$, and $z$ simultaneously produces better results than first determining $x$, $y$ and then interpolating $z$ linearly. It suggests that $z$ depends not only on $x$ and $y$, but also vice versa. 
M3 achieves a median error less than half that of the linear model, which highlights its effectiveness.

\vspace*{10mm}

However, we also observe that M2 and M3 produce more outliers than M1. In some cases, the error of M2 exceeds even the largest error of M1. This is likely due to the differing target distributions between the linear and nonlinear models.
The neural networks in M2 and M3 were trained on data from M1, which did not include edge points. However, the target points for M2 and M3 do include edge cases, which requires extrapolation beyond the training data, resulting in larger errors.

\begin{figure}[b]
\centering
\includegraphics[width=0.8\textwidth]{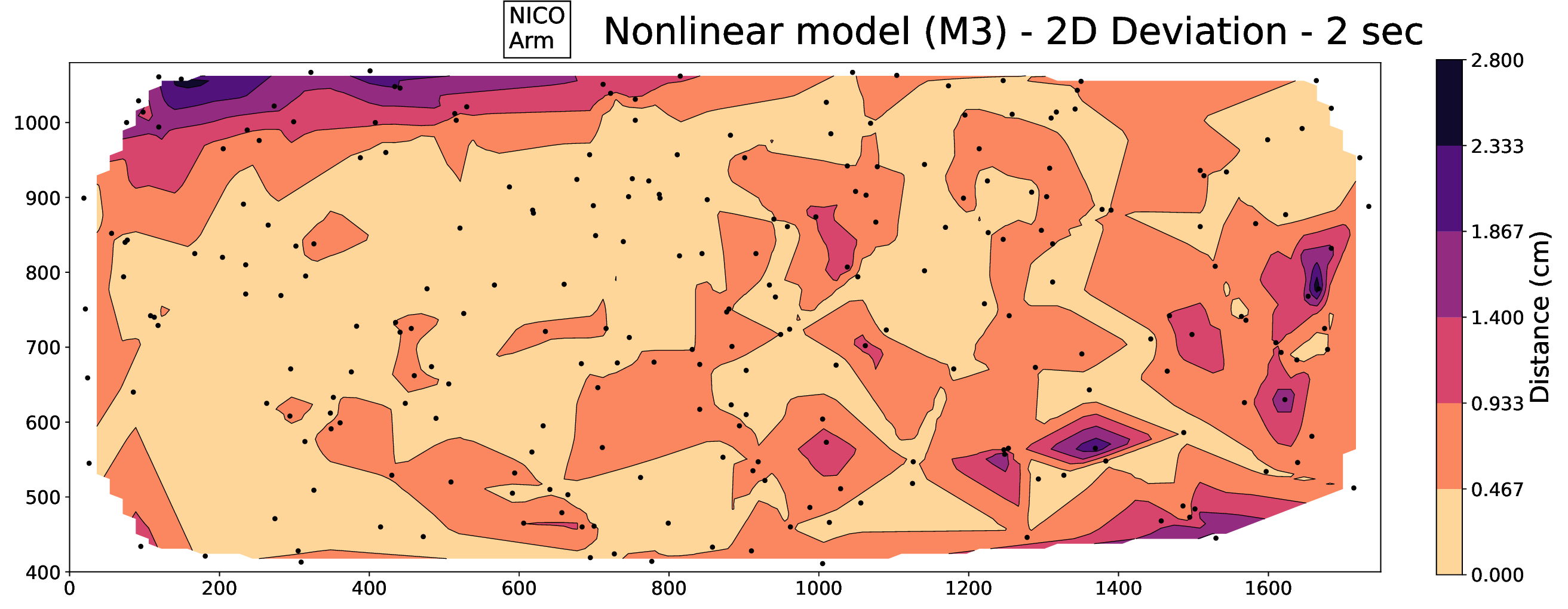}
\caption{Accuracy of NICO robot after the neural network based correction, trained to predict 3D points in the simulator space from target 2D points on the touchscreen. Accuracy is calculated as a deviation in 2D space.} 
\label{m3_contour}
\end{figure}

We identified the quadrants where our nonlinear model produces the highest errors, but we were also interested in pinpointing the exact locations of the outliers seen in Fig.~\ref{models_boxplot}.
For this, we visualized the accuracy of the model in the entire touchscreen testing area (Fig.~\ref{m3_contour}).
The visualization confirms that the largest errors occur in Q4, where the EE requires the longest trajectory. Furthermore, smaller regions in Q2 also exhibit high errors, particularly at extreme points that require extrapolation. 
In contrast, Q3 consistently shows the best results.

\begin{table}[t]
\caption{Accuracy of NICO robot 2-sec movement in 4 quadrants using our three models. Each cell contains the mean error and the standard deviation in centimeters.}
\label{models_table}
\resizebox{\textwidth}{!}{
\begin{tabular}{lccccc}
\hline
Model &  Q1 & Q2 & Q3 & Q4 & Mean\\
\hline
M1 & $0.78 \pm 0.45$ & $1.00 \pm {0.27}$ & $1.46 \pm 0.53$ & $1.34 \pm 0.66$ & $1.15 \pm 0.58$ \\
M2 &  $0.76 \pm 0.45$ & $1.18 \pm 0.88$ & $0.62 \pm {0.29}$ & $0.80 \pm {0.39}$ & $0.88 \pm 0.65$ \\
M3 &  $\mathbf{0.56} \pm \mathbf{0.42}$ & $\mathbf{0.61} \pm \mathbf{0.53}$ & $\mathbf{0.46} \pm \mathbf{0.29}$ & $\mathbf{0.70} \pm \mathbf{0.53}$ & $\mathbf{0.58} \pm \mathbf{0.46}$ \\
\hline
\end{tabular}
}
\end{table}

Table~\ref{models_table} presents the accuracy of our models in all four quadrants, plus the mean accuracy. The data set for M1 contains 130 regularly distributed points in certain patterns (e.g., grid) to visualise the direction of deviations and 120 randomly distributed points over the testing area of the touchscreen. The data set for M2 and M3 contains 250 random points, including points near the edge of the touchscreen. The data set for M1 does not include edge points because for many of them, NICO would hit outside the screen, making them unusable.
In most cases, the error decreases as the complexity of the model increases. However, in Q2, M2 showed a slight increase in error from 1 cm to 1.18 cm, and also a significant increase in standard deviation, from 0.27 to 0.88 cm. This correlates with M2 having the most outliers in Fig.~\ref{models_boxplot}.
For M3, Q3 showed the lowest average error, while Q4, being farthest from the base, showed the opposite. Overall, when considering all quadrants together, the mean error was reduced from 1.15 cm in the linear model to 0.58 cm in M3. Furthermore, the standard deviation decreased, indicating greater stability in our solution.

\vspace*{-2mm}
\section{Discussion}
\vspace*{-1mm}

Robotic tasks are highly specialized, and each study uses different testing methodologies, making direct comparisons challenging. In the work most comparable to ours, the same NICO robot was used to grasp objects \cite{gade2024}. Their best-performing model achieved an 80.3\% grasp success rate with an accuracy of 0.93 cm. This was accomplished through a combination of full domain randomization, introducing variability in camera angles and visual features, and random-to-canonical adaptation, which enhanced sim-to-real transfer and outperformed a real-world-trained model that used six times more data. In contrast, our model, without relying on any visual input and using only a touchscreen and a neural network, achieved a higher accuracy of 0.58 cm, although the task was different.
Roncone et al. \cite{roncone2014}, mentioned in Sec.~2, achieved a positional accuracy of 0.21 cm with the iCub humanoid robot using an automated self-touch calibration method. 
Although they report lower errors than ours, they use expensive visual systems and artificial skin, whereas our approach relies solely on a touchscreen and a simple neural network. And more importantly, iCub is an order of magnitude more expensive robot than NICO, which entails hardware properties.

This work focuses on haptic feedback in the 2D horizontal plane (touchpad), which may be seen as a limitation. However, even the 2D case turned out to be rather complex because of its nonlinearity. Extension to 3D is a natural next step, posing its own challenges (e.g. how to provide 3D haptic feedback).

In summary, we presented a novel calibration method to improve sim-to-real transfer in robotic manipulation using haptic feedback. Using a touchscreen to obtain the robot's end-effector positions, we developed transformation models to effectively bridge the discrepancy between simulation and reality. Our results demonstrate that a fully nonlinear neural network model significantly reduces positioning errors, outperforming traditional linear approaches. This low-cost and adaptable method can be applied to various robotic systems. 

\begin{credits}
\subsubsection{\ackname}  This research was supported by the Horizon Europe TERAIS project, no. 101079338 and, partially, by the national project VEGA 1/0373/23 (IF).
\end{credits}
%
%
%

%


\begin{thebibliography}{10}
\providecommand{\url}[1]{\texttt{#1}}
\providecommand{\urlprefix}{URL }
\providecommand{\doi}[1]{https://doi.org/#1}

\bibitem{alquareshi2020}
Al-Qurashi, Z.: Robot-human mapping model learning for robotic imitation using
  deep learning and virtual reality. Ph.D. thesis, Univ. of Illinois at Chicago
  (2020)

\bibitem{coumans2021}
Coumans, E., Bai, Y.: Py{B}ullet, a python module for physics simulation for
  games, robotics and machine learning. \url{http://pybullet.org} (2016--2021)

\bibitem{gade2024}
G{\" a}de, C., Habekost, J.G., Wermter, S.: Domain adaption as auxiliary task
  for sim-to-real transfer in vision-based neuro-robotic control. In: IJCNN.
  IEEE (2024)

\bibitem{hogan2022}
Hogan, F., Carlo, J.D., Kim, S.: Proprioceptive control for dynamic quadrupedal
  locomotion. Science Robotics  \textbf{7}(62),  eabk2822 (2022)

\bibitem{hubert2012}
Hubert, U., St{\"u}ckler, J., Behnke, S.: Bayesian calibration of the hand-eye
  kinematics of an anthropomorphic robot. In: Int. Conf. on Humanoid Robots
  (2012)

\bibitem{kappassov2015}
Kappassov, Z., Hostettler, J.B., Billard, A.: Tactile sensing in dexterous
  robot hands – review. Robotics and Autonomous Systems  \textbf{74},
  195--220 (2015)

\bibitem{nico_software}
Kerzel, M., Strahl, E., Magg, S., Navarro-Guerrero, N., Heinrich, S., Wermter,
  S.: {NICO} - {N}euro-{I}nspired {CO}mpanion: A developmental humanoid robot
  platform for multimodal interaction. In: IEEE RO-MAN. pp. 113--120 (2017)

\bibitem{khusainov2017}
Khusainov, R., Klimchik, A., Magid, E.: Humanoid robot kinematic calibration
  using industrial manipulator. In: ICMSC. pp. 184--189. IEEE (2017)

\bibitem{levine2018}
Levine, S., Abbeel, P., Finn, C., Darrell, T.: Learning hand-eye coordination
  for robotic grasping with deep learning and large-scale data collection. The
  International Journal of Robotics Research  \textbf{37}(4-5),  421--436
  (2018)

\bibitem{liu2022}
Liu, W., Zhao, C., Liu, Y., Wang, H., Zhao, W., Zhang, H.: Sim2real kinematics
  modeling of industrial robots based on {FPGA}-acceleration. Robotics and
  Computer-Integrated Manufacturing  \textbf{77} (2022)

\bibitem{hwei2017nico}
Ng, H.G., et~al.: Hey robot, why don't you talk to me? In: IEEE International
  Symposium on Robot and Human Interactive Communication (RO-MAN) (2017)

\bibitem{nguyen2018}
Nguyen, P.D., Fischer, T., Chang, H.J., Pattacini, U., Metta, G., Demiris, Y.:
  Transferring visuomotor learning from simulation to the real world for
  robotics manipulation tasks. In: IEEE IROS. pp. 6667--6674 (2018)

\bibitem{vincente2014}
Pedro, V.: Real time graphical simulation for visual based pose estimation and
  self-calibrating of a humanoid robotic arm. Master's thesis, Instituto
  Superior Técnico, Lisbon (2014)

\bibitem{roncone2014}
Roncone, A., Hoffmann, M., Pattacini, U., Metta, G.: Automatic kinematic chain
  calibration using artificial skin: self-touch in the i{C}ub humanoid robot.
  In: ICRA. pp. 2305--2312. IEEE (2014)

\bibitem{rozlivek2021}
Rozlivek, J., Rustler, L., \v{S}tep\'anov\'a, K., Hoffmann, M.: Multisensorial
  robot calibration framework and toolbox. In: Int. Conf. on Humanoid Robots.
  IEEE (2021)

\bibitem{tenhumberg2021}
Tenhumberg, J., B{\"a}uml, B.: Calibration of an elastic humanoid upper body
  and efficient compensation for motion planning. In: Int. Conf. on Humanoid
  Robots. IEEE (2021)

\bibitem{vavrecka2021}
Vavre\v{c}ka, M., Sokovnin, N., Mejdrechov\'a, M., \v{S}ejnov\'a, G.: My{G}ym:
  Modular toolkit for visuomotor robotic tasks. In: ICTAI. pp. 279--283 (2021)

\bibitem{stepanova2019}
\v{S}tep\'anov\'a, K., Pajdla, T., Hoffmann, M.: Robot self-calibration using
  multiple kinematic chains—a simulation study on the i{C}ub humanoid robot.
  IEEE Robotics and Automation Letters  \textbf{4}(2),  1900--1907 (2019)

\end{thebibliography}
\end{document}